\pdfoutput=1

\documentclass[11pt]{article}

\usepackage{ACL2023}

\usepackage{times}
\usepackage{latexsym}

\usepackage[T1]{fontenc}

\usepackage[utf8]{inputenc}

\usepackage{microtype}

\usepackage{inconsolata}

\usepackage[pdftex]{graphicx}

\newcommand{\yn}[1]{\textcolor{black}{#1}}

\makeatletter
\newcommand\footnoteref[1]{\protected@xdef\@thefnmark{\ref{#1}}\@footnotemark}
\makeatother

\newcommand{\pluseq}{\mathrel{+}=}

\title{Back to 
Patterns:\\
Efficient Japanese Morphological Analysis with Feature-Sequence Trie}

\author{Naoki Yoshinaga \\
  Institute of Industrial Science, The University of Tokyo \\
  \texttt{ynaga@iis.u-tokyo.ac.jp}\\}

\usepackage[ruled]{algorithm}
\usepackage[noend]{algorithmic}

\usepackage{bm}
\usepackage{amsmath}
\usepackage{amssymb}
\usepackage{booktabs}

\DeclareMathOperator*{\argmax}{argmax}

\begin{document}
\maketitle
\begin{abstract}
Accurate neural models are much less efficient than non-neural models and are useless for processing billions of social media posts or handling 
user queries in real time with a limited budget.
This study revisits the fastest pattern-based \textsc{nlp} methods to make them as accurate as possible, thus yielding a strikingly
simple yet surprisingly accurate morphological analyzer for Japanese.
The proposed method induces reliable patterns from a morphological dictionary and annotated data. Experimental results on two standard datasets confirm that the  method exhibits comparable accuracy to learning-based baselines, while boasting a remarkable throughput of \textbf{over 1,000,000 sentences per second} on a single modern \textsc{cpu}\@.
The source code is available at \url{https://www.tkl.iis.u-tokyo.ac.jp/~ynaga/jagger/}.
\end{abstract} 

\section{Introduction}
The amount of text data being processed has greatly increased since the advent of communication platforms such as Twitter, Zoom, and Slack, and \textsc{nlp} services such as DeepL and Grammarly
have millions of users. Some users analyze textual big data for marketing, linguistics, or sociology, while others deploy \textsc{nlp} services on their own devices because of privacy concerns.
It is therefore becoming important to develop highly efficient methods to
process massive text data and user queries with limited computational resources.

However, the recent campaign for efficient \textsc{nlp} does not focus on literally efficient methods that scale to increasing data sizes and run on resource-constrained devices. 
Instead, most ``efficient'' \textsc{nlp} studies~\citep{arxiv:2209.00099} focus on neural methods, which \yn{are too slow to handle
billions of social media posts and too large to deploy on edge devices.} Those studies seek to make model training or inference \textit{relatively} efficient within the deep learning framework. Thus, the large efficiency gap 
with respect to classical methods has never been filled. 



\begin{figure}
    \centering
    \includegraphics[width=\linewidth,clip]{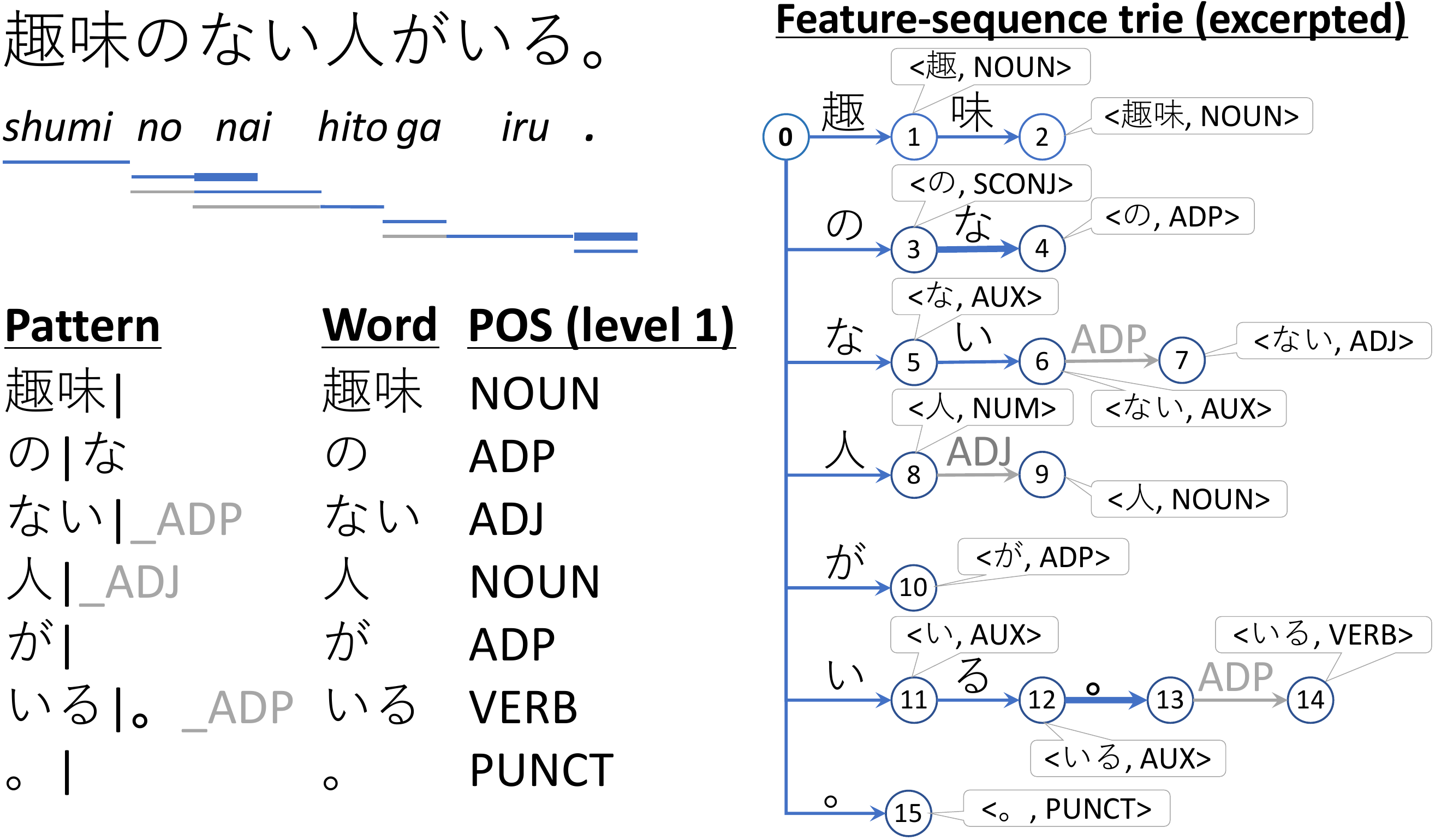}
    \caption{Pattern-based morphological analysis via a feature-sequence trie. The \textcolor{blue}{blue} and \textcolor{gray}{gray} lines below the input indicate pattern matches (trailing characters and previous \textsc{pos} tags) to determine where to split (indicated by `|' in the patterns) and what to tag.} 
    \label{fig:example}
\end{figure}

In this study, I take an orthogonal approach toward \textit{absolutely} efficient \textsc{nlp} by seeking to boost the accuracy of the fastest methods. Specifically, I
have developed a remarkably simple yet accurate method for Japanese morphological analysis, which is a joint task of word segmentation,
part-of-speech (\textsc{pos}) tagging, and lemmatization. This method \yn{revisits} the classical longest matching method;  it greedily applies patterns that determine the next position to segment and then identifies the \textsc{pos}
tag for the segmented word, as illustrated in Figure~\ref{fig:example}. To obtain reliable patterns, starting from words in a morphological dictionary and training data, patterns are extended with posterior surface contexts and previous \textsc{pos} tags, and the patterns' segmentation offsets and tags are determined by frequency. The extracted patterns are then stored in an efficient double-array trie~\citep{aoe1989}.


The proposed method was evaluated on two standard corpora~\citep{kurohashi2003,hangyo-etal-2012-building}.
The experimental results confirmed that this simple method can process 1,000,000 sentences per second on an M2 MacBook Air, with comparable accuracy to learning-based baselines~\citep{kudo-etal-2004-applying,neubig-etal-2011-pointwise}.

\section{Pattern-based Morphological Analysis}
This section describes the method of Japanese morphological analysis used here, which performs word segmentation, \textsc{pos} tagging, and lemmatization. To maximize the tagging efficiency, I return to a pattern-based algorithm that is similar to the longest matching algorithm~\citep{nagata-1994-stochastic}. 

The longest matching algorithm performs deterministic word segmentation by using a dictionary. Starting from the beginning of the input, it greedily finds the longest dictionary words to segment the input.
Although this \yn{simple}
algorithm exhibits moderate \yn{accuracy} in Chinese and Japanese with
transformation rules~\citep{palmer-1997-trainable,hockenmaier-brew-1998-error,sassano-2014-deterministic}, there is a gap in accuracy
from search- and classification-based approaches~\citep{kudo-etal-2004-applying,neubig-etal-2011-pointwise}. \yn{To make search-based morphological analysis partially deterministic, \citet{morita-iwakura-2019-fast} extracted surface patterns from tagging results; however, the speed-up factor was at most 1.5.}


\begin{algorithm}[t]
  \small
  \caption{Pattern-based morphological analysis}\label{algo:gm}
  \begin{algorithmic}[1]
    \REQUIRE sequence of characters, $\bm{c}$; set of patterns stored in trie, $\mathcal{P} = \{(\textsf{p}, \textsf{shift}, t)\}$ 
    \ENSURE 
    sequence of words with tags $\bm{s} = \{(w_j, t_j)\}$
    \STATE $i \leftarrow 0$
    \WHILE{$i < \texttt{len} (\bm{c})$}
        \STATE \mbox{$(\hat{\textsf{shift}}, \hat{t}) = \texttt{longest\_prefix\_search}(\bm{c}_{\ge i}, \mathcal{P})$}
        \STATE $\texttt{append}(\bm{s}, (\bm{c}_i^{i + \hat{\textsf{shift}}}, \hat{t}))$
        \STATE $i \leftarrow i + \hat{\textsf{shift}}$
    \ENDWHILE
    \RETURN $\bm{s}$
  \end{algorithmic}
\end{algorithm}

\subsection{Basic algorithm}
Algorithm~\ref{algo:gm} is a simple, deterministic algorithm for joint word segmentation, \textsc{pos} tagging, and lemmatization. It repeatedly applies the longest-matching patterns in a trie $\mathcal{P}$ to a given sequence of characters, $\bm{c}$, and a start position $i$ to segment and tag the next word ($w_j=\bm{c}_i^{i+\hat{\textsf{shift}}}$ and $\hat{t}_j$). As will be shown later in \S~\ref{sec:experiments}, this simple algorithm \textit{works} as well as learning-based approaches.


\yn{This algorithm is inspired by the longest matching algorithm but differs in that the segmentation offset \textsf{shift} can be smaller than the surface length matched with patterns, $k$ (see Line~\ref{l:k} in Algorithm~\ref{algo:pat}). A running example is shown in Figure~\ref{fig:example}.}

The algorithm is also inspired by the precomputation of feature weights in sequence labeling~\citep{kaji-etal-2010-efficient} and classification with conjunctive features~\citep{yoshinaga-kitsuregawa-2009-polynomial,yoshinaga-kitsuregawa-2010-kernel,yoshinaga-kitsuregawa-2014-self}.
Those methods accumulate certain feature weights in advance and retrieve those partial results by using \yn{simple keys such as word unigrams, \textsc{pos} bigrams, and primitive feature sequences to compute the final results (labels) by an $\argmax$ operation on the weights.
The proposed method regards word segmentation and tagging as a joint, multi-class classification problem and directly obtains the label
(i.e., where to segment and what to tag) by using the feature sequence as a pattern, thus skipping the expensive $\argmax$ operation over a number of labels. The longest matching thus 
implies classification with as many features as possible.}

\begin{algorithm}[t]
  \small
  \caption{Pattern extraction from training data}\label{algo:pat}
  \begin{algorithmic}[1]
    \REQUIRE training data $\mathcal{D}$ and dictionary $\mathcal{V}$
    \ENSURE set of patterns, $\mathcal{P} = \{(p, \textsf{shift}, t)\}$
    \STATE $\hat{\mathcal{P}}\leftarrow \phi$
    \STATE $L_{\textrm{max}} = \max_{(w, t) \in\mathcal{V}} \texttt{len}(w)$
    \FORALL{training examples $(\bm{c}, \bm{s}=\{(w_l, t_l)\}_{l=1}^{L}) \in \mathcal{D}$}\label{l:es}
        \STATE $i\leftarrow 0$
        \FOR{$j = 0$ to $L$}
            \STATE $\textsf{shift} = \texttt{len}(w_j)$
            \FOR{$k = \textsf{shift}$ to $L_{\textrm{max}}$}\label{l:k}
                \STATE $\hat{\mathcal{P}}[\bm{c}_i^{i+k}][(\textsf{shift}, t_j)]\pluseq 1$
                \STATE $\hat{\mathcal{P}}[\bm{c}_i^{i+k};t_{j-1}][(\textsf{shift}, t_j)]\pluseq 1$
            \ENDFOR
            \STATE $i \leftarrow i + \textsf{shift}$\label{l:ee}
        \ENDFOR
    \ENDFOR
    \STATE $\mathcal{P} \leftarrow \{(w, \texttt{len}(w), \hat{t})\}$ where $(w, *) \in \mathcal{V}, w \not\in \hat{\mathcal{P}},$\label{l:dicts}
    \STATE\hspace{\algorithmicindent} $\hat{t} = \argmax_{\{t \mid (w,t)\in\mathcal{V}\}} \sum_{w'} \hat{\mathcal{P}}[w'][(\texttt{len}(w'), t)]$\label{l:dicte}
    \FORALL{pattern candidates $p \in \hat{\mathcal{P}}$ from shortest one}
        \STATE $\textsf{shift} = \argmax_{\textsf{shift}} \sum_{t}\hat{\mathcal{P}}[p][(\textsf{shift}, t)]$\label{l:shift}
        \STATE $t = \argmax_{t} \hat{\mathcal{P}}[p][\textsf{shift}, t)]$\label{l:tag}
        \STATE $(\textsf{shift}', t') = \texttt{longest\_prefix\_search}(p, \mathcal{P})$\label{l:prunes}
        \IF{$(\textsf{shift}, t) = (\textsf{shift}', t')$}
            \STATE $\mathcal{P} \leftarrow \mathcal{P} \cup \{(p, \textsf{shift}, t)\}$\label{l:prunee}
        \ENDIF
    \ENDFOR
    \RETURN $\mathcal{P}$
  \end{algorithmic}
\end{algorithm}

\subsection{Pattern extraction from data}
Following the feature templates of learning-based methods~\citep{kudo-etal-2004-applying,neubig-etal-2011-pointwise}, the algorithm's pattern template was designed as a sequence of characters, $\bm{c}$, followed by the previous word's \textsc{pos} tag $t_{j-1}$, thus giving $\bm{c};t_{j-1}$, where `$;$' represents string concatenation.

Algorithm~\ref{algo:pat} is the procedure to extract patterns for word segmentation and \textsc{pos} tagging from the annotated data and a dictionary. Given training data $\mathcal{D}$ with annotation of (word) segmentations and (\textsc{pos}) tags and a dictionary $\mathcal{V}$ compiling words and their possible tags, the algorithm iteratively extracts possible patterns from $\mathcal{D}$. It first enumerates surface patterns $\bm{c}_i^{i+k}$ from all starting positions of words in $\mathcal{D}$, and it then concatenates them with tag $t_{j-1}$ for the preceding words to form pattern candidates (Lines~\ref{l:es}-\ref{l:ee} in Algorithm~\ref{algo:pat}). Patterns are added for dictionary words that are unseen in the training data (Lines~\ref{l:dicts}-\ref{l:dicte}). The segmentation offset ($\textsf{shift}$) and tag $t$ for a pattern are determined by the frequency (Lines~\ref{l:shift}-\ref{l:tag}). 
\yn{To avoid extra matching to the posterior contexts and previous tag,
we only keep patterns whose segmentation offsets and tags differ from those of the longest \textit{prefix} patterns that share prefixes of posterior contexts (Lines~\ref{l:prunes}-\ref{l:prunee}). 
This not only reduces the number and length of patterns but also minimizes the longest matching method's overhead for word segmentation.\footnote{In preliminary experiments, a variant of backtracking-free search~\citep{maruyama-1994-backtracking} did not improve the throughput.}}




\section{Experiments}
\label{sec:experiments}
This section describes an experimental evaluation of the pattern-based morphological analyzer on two annotated corpora in different domains~\citep{kurohashi2003,hangyo-etal-2012-building}. The method was compared with two learning-based baselines~\citep{kudo-etal-2004-applying,neubig-etal-2011-pointwise} in terms of efficiency and accuracy. Note that all language resources and software used in the experiments are publicly available and free for academic use.

\subsection{Setup}\label{ssec:setup}
\paragraph{Data} The experiments used the Kyoto-University Text Corpus\footnote{\url{https://github.com/ku-nlp/KyotoCorpus}} (\textsc{kyoto})~\citep{kurohashi2003}, compiled from newspaper articles, and the Kyoto-University Web Document Leads Corpus\footnote{\url{https://github.com/ku-nlp/KWDLC}} (\textsc{kwdlc})~\citep{hangyo-etal-2012-building}, compiled from the first three sentences of various Web pages. I adopted the split of development and test sets given in the corpora's \texttt{github} repositories and used the remaining portions as training sets. The datasets' statistics are listed in Table~\ref{tab:corpus}.

\begin{table}[t]
\centering
\small
\tabcolsep3pt
\begin{tabular}{ccccccc}
\toprule
 & \multicolumn{3}{c}{\textsc{kyoto}} & \multicolumn{3}{c}{\textsc{kwdlc}}  \\
 \cmidrule(lr){2-4}
 \cmidrule(lr){5-7}
 & train & dev & test & train & dev & test \\
\midrule
\# sentences & 35,478 & 1145 & 1783 & 12,271 & 1585 & 2195 \\
ave. \# words &  25.37 & 26.24 & 25.83 & 15.85 & 14.27 & 16.34 \\
\bottomrule
\end{tabular}
\caption{Statistics of the evaluation datasets.}\label{tab:corpus}
\end{table}

\paragraph{Methods} The three methods below were compared. \yn{To prevent overfitting,
the hyperparameter $C$ in the underlying model was tuned for the two learning-based 
baseline methods\footnote{$C=\{0.1, 0.2, 0.5, 1.0, 2.0, 5.0, 10.0\}$.}} by using the development set to maximize the F$_1$ of the \textsc{pos} tags.


\smallskip\noindent\textbf{MeCab} (ver.~0.996) is a C++ implementation of a search-based 
method~\citep{kudo-etal-2004-applying}.\footnote{\url{https://taku910.github.io/mecab/}} It enumerates possible segmentations and tags as word lattices by using a dictionary and performs Viterbi search by using unigram and bigram scores factorized from feature weights.


\begin{table}[t]
\centering
\small
\begin{tabular}{lrrrrrr}
\toprule
& \# words & \multicolumn{5}{c}{\# tags (four levels)} \\
\cmidrule(lr){3-7}
& & 1 & 2 & 3 & 4 & all (1-4)\\
\midrule
\textsc{juman} 5.1 & 475,716 & 14 & 35 & 34 & 60 & 980 \\
\textsc{juman} 7.0 & 702,358 & 14 & 35 & 33 & 77 & 1,188 \\
\bottomrule
\end{tabular}
\caption{Statistics of the morphological dictionaries.}\label{tab:dict}
\end{table}

\smallskip\noindent\textbf{Vaporetto} (ver.~0.6.2) is a Rust\footnote{Rust exhibits comparable efficiency to C++ on program benchmarks: \url{https://github.com/kostya/benchmarks/}.} implementation of a classification-based method~\citep{neubig-etal-2011-pointwise}.\footnote{\url{https://github.com/daac-tools/vaporetto}} It first performs word segmentation by classifying \yn{whether to segment} 
after each character in the input, and it then identifies the resulting words' \textsc{pos} tags.
It also trains classifiers for the possible \textsc{pos} tag sets of individual words, and it assigns the \textsc{pos}s of its first dictionary entries for words that are unseen
in the training data.\footnote{Words that did not appear in the dictionary were assigned ``SAHEN noun,'' following \citet{kudo-etal-2004-applying}. The efficiency results below do not include this postprocessing.} A morphological dictionary was used to extract word features.

\smallskip\noindent\textbf{Jagger} is a C++ implementation of the proposed algorithm.
\yn{It greedily applies patterns extracted from the training data and a dictionary to jointly segment words and assign tags.} 
Appendices~\ref{app:unk} and~\ref{app:impl} respectively describe the method to handle unknown words and the implementation details.
\yn{Jagger is more similar to Vaporetto than to MeCab
but differs in that it jointly performs segmentation and tagging instead of using a two-step cascaded pipeline, and it uses patterns instead of classifiers to
find labels (i.e., where to segment and what to tag).} Appendix~\ref{app:vib} compares Jagger with the other implementations.


\begin{table*}[t]
\centering
\small
\tabcolsep3.8pt
\begin{tabular}{l@{\,}r@{\,\,}r@{\,\,}r@{\qquad}ccc}
\toprule
\textsc{kyoto} & time {\scriptsize [s] $\downarrow$} & speed {\scriptsize [sent./s] $\uparrow$} & space {\scriptsize [MiB]  $\downarrow$} & seg & top (level 1) & all (levels 1-4)\\
\midrule
\multicolumn{7}{c}{w/ jumandic-5.1} \\
 MeCab       & 26.83 & 66,455 & 55.81 & 98.68 (98.47/98.89) & 97.32 (97.12/97.53) & 95.97 (95.76/96.17)\\
Vaporetto   & 
15.14  & 117,767 & 658.80 &
98.94 (98.97/98.92) & 98.30 (98.32/98.27) & 96.92 (96.95/96.90) \\
Jagger (proposed) &  1.77 & 1,007,344 & 26.39 & 98.73 (98.62/98.83) & 97.62 (97.52/97.72) & 96.55 (96.45/96.65) \\
\midrule
\multicolumn{7}{c}{w/ jumandic-7.0} \\
MeCab       & 29.99 & 59,453 & 77.98 & 98.37 (98.02/98.72) & 97.19 (96.84/97.54) & 96.10 (95.75/96.44) \\
Vaporetto   & 
16.93 & 105,316 &  828.85 &
99.08 (99.08/99.08) & 98.42 (98.42/98.43) & 97.05 (97.04/97.05)  \\
Jagger (proposed) &  1.83 & 974,316 & 35.09 & 98.68 (98.51/98.86) & 97.63 (97.46/97.80) & 96.57 (96.74/96.40) \\
\bottomrule
\end{tabular}
\caption{F$_1$ (precision/recall) results on \textsc{kyoto}.}\label{tab:kc}
\end{table*}

\begin{table*}[t]
\centering
\small
\tabcolsep3.8pt
\begin{tabular}{l@{\,}r@{\,\,}r@{\,\,}r@{\qquad}ccc}
\toprule
\textsc{kwdlc} & time {\scriptsize [s] $\downarrow$} & speed {\scriptsize [sent./s] $\uparrow$} & space {\scriptsize [MiB]  $\downarrow$} & seg & top (level 1) & all (levels 1-4)\\
\midrule
\multicolumn{7}{c}{w/ jumandic-5.1} \\
MeCab       &  23.83 & 92,110 & 53.88 & 97.13 (96.82/97.44) & 95.62 (95.32/95.93) & 94.30 (94.00/94.60) \\
Vaporetto   &  10.93  &  200,823 & 642.63
& 97.35 (97.39/97.32) & 96.16 (96.20/96.13) & 94.08 (94.11/94.04) \\
Jagger (proposed) & 1.44 & 1,524,305 & 28.89 & 97.17 (96.94/97.40) & 95.71 (95.49/95.94) & 94.20 (93.98/94.42)\\
\midrule
\multicolumn{7}{c}{w/ jumandic-7.0} \\
MeCab       & 26.90 & 81,598 & 76.38 & 97.99 (97.82/98.16) & 96.66 (96.49/96.83) & 95.62 (95.45/95.78)\\
Vaporetto   & 12.55  & 174,900  & 842.40
& 97.53 (97.58/97.49) & 96.39 (96.43/96.34) & 94.68 (94.72/94.63) \\
Jagger (proposed) & 1.46 & 1,503,424 & 40.22 & 97.60 (97.49/97.71) & 96.14 (96.04/96.25) & 94.63 (94.52/94.73)\\
\bottomrule
\end{tabular}
\caption{F$_1$ (precision/recall) results on \textsc{kwdlc}.}\label{tab:kwdlc}
\end{table*}

\paragraph{Dictionaries}
As listed in Table~\ref{tab:dict}, the experiments used two morphological dictionaries imported to MeCab from a manually tailored morphological analyzer, \textsc{juman}.\footnote{\url{https://nlp.ist.i.kyoto-u.ac.jp/?JUMAN}}
Specifically, mecab-jumandic-5.1-20070304 and mecab-jumandic-7.0-20130310 were compared to examine the impact of the dictionary's quality and size.
The jumandic-7.0 dictionary contains words extracted automatically from the Web~\citep{murawaki-kurohashi-2008-online}, comprising a larger number (702,358) than in jumandic-5.0 (475,716). The \textsc{pos} tags include four levels of hierarchical morphosyntactic information: (1) major \textsc{pos} (\textit{e.g.}, \textit{noun} and \textit{verb}); (2) minor \textsc{pos} (\textit{e.g.}, \textit{common noun}); (3) conjugation type (\textit{e.g.}, \textit{ichidan verb}); and (4) conjugation form (\textit{e.g.}, \yn{\textit{irrealis}}). \yn{For example, the \textsc{pos} tags of \textit{shumi} and \textit{iru} in Figure~\ref{fig:example} are \textit{noun}-\textit{common\_noun}-*-* and \textit{verb}-*-\textit{ichidan\_verb}-\textit{terminal}, respectively.}

\paragraph{Evaluation procedure} The precision, recall, and F$_1$ of the segmentation with various levels of \textsc{pos} tags~\citep{kudo-etal-2004-applying} were used as metrics. \yn{As Vaporetto does not output lemmas, lemmatization was evaluated via the tagging results of the full \textsc{pos} tag set (``all (levels 1-4)'' in Tables~\ref{tab:kc} and~\ref{tab:kwdlc}), which included conjugation types and forms, given that Japanese words can be mapped to their lemmas according to their conjugation types and forms.} I processed 1000 copies of the test data and measured the 
time, 
speed, and maximum memory consumption three times with the \texttt{/usr/bin/time -l} command. The median values are reported here. All experiments were done on an M2 MacBook Air with a 3.5-GHz CPU and 24-GB main memory. 


\begin{table*}[t]
\tabcolsep4.2pt
\centering
\small
\tabcolsep3.8pt
\begin{tabular}{l@{\,}r@{\,\,}r@{\,\,}r@{\qquad}ccc}
\toprule
& time {\scriptsize [s] $\downarrow$} & speed {\scriptsize [sent./s] $\uparrow$} & space {\scriptsize [MiB] $\downarrow$} & seg & top (level 1)  & all (levels 1-4)\\
\midrule
\multicolumn{7}{c}{\textsc{kyoto}} \\
\textsc{juman++-v2} & 331.14 & 5384 & 300.80 & 99.37 (99.30/99.45) & 98.72 (98.65/98.80) & 97.74 (97.66/97.82) \\
Jagger (proposed) &  1.83 & 974,316 & 35.09 & 98.68 (98.51/98.86) & 97.63 (97.46/97.80) & 96.57 (96.74/96.40) \\
\midrule
\multicolumn{7}{c}{\textsc{kwdlc}} \\
\textsc{juman++-v2} & 283.11 & 7753 & 290.05 & 98.37 (98.25/98.50) & 97.61 (97.49/97.73) & 96.42 (96.30/96.55) \\
Jagger (proposed) &  1.46 & 1,503,424 & 40.22 & 97.60 (97.49/97.71) & 96.14 (96.04/96.25) & 94.63 (94.52/94.73)\\
\bottomrule
\end{tabular}
\caption{F$_1$ (precision/recall) comparison with \textsc{juman}++.}\label{tab:neural} 
\end{table*}

\subsection{Results}
Tables~\ref{tab:kc} and \ref{tab:kwdlc} summarize the \yn{morphological analysis} results on the \textsc{kyoto} and \textsc{kwdlc} datasets. The pattern-based method here, Jagger, was 16 and 7 times faster than MeCab and Vaporetto with 1/2 and 1/20 as much memory consumption, respectively, while achieving comparable accuracy. Jagger is efficient because it does not have massive floating-point parameters, unlike other methods, and because it minimizes the number and length of patterns by pruning (Lines~\ref{l:prunes}-\ref{l:prunee} in Algorithm~\ref{algo:pat}). As a result, the training took less than six seconds. MeCab's accuracy 
depends on the dictionary: with jumandic-7.0, it worked best on \textsc{kwdlc} and worst on \textsc{kyoto}. In contrast, Vaporetto's accuracy 
depends on the training data size. It worked best on \textsc{kyoto} but was just as good as Jagger on \textsc{kwdlc}.

Below are the detailed results for Jagger with the jumandic-7.0 dictionary.


\paragraph{Comparison to neural methods}
Jagger was compared to a state-of-the-art neural method~\citep{tolmachev-etal-2018-juman}, \textsc{juman++-v2},\footnote{\url{https://github.com/ku-nlp/jumanpp}} which was trained on the same data with the official script and hyperparameters.\footnote{\url{https://github.com/ku-nlp/jumanpp-jumandic}} Note that this comparison was \textbf{unfair} to Jagger in terms of accuracy and to \textsc{juman++-v2} in terms of efficiency, because \textsc{juman++-v2} uses 0.8 million additional dictionary entries from Wikipedia and a neural language model trained on 10 million sentences from the Web.

Table~\ref{tab:neural} summarizes the comparison between Jagger and \textsc{juman++-v2}. Although \textsc{juman++-v2} was reported to speed up \textsc{juman}++~\citep{morita-etal-2015-morphological} by a factor of 250, Jagger was faster than \textsc{juman++-v2} by a factor of 180 with 1/7 as much of a memory footprint. \textsc{juman++-v2} was more accurate than Jagger, but the gain was less than 1\% for word segmentation. If external text could be used, this gap could be reduced 
with a technique called structure compilation~\citep{10.1145/1390156.1390231}, which runs \textsc{juman++-v2} on external text to extract patterns. That idea is beyond this paper's scope but important for future work.


\paragraph{Word segmentation efficiency} Because of different approaches to handling unknown words and supporting lemmatization, it is difficult to compare Vaporetto with Jagger and MeCab as a morphological analyzer in a strictly fair manner. Instead, the word segmentation efficiency was compared, as summarized in Table~\ref{tab:seg}. 
Here, Vaporetto was trained to perform only word segmentation by using \yn{the dictionary and the training data without \textsc{pos} tags.} Jagger was faster and more space-efficient than Vaporetto, even taking the overhead of loading large models (1.7 seconds) into account.

\begin{table}[t]
\centering
\small
\tabcolsep3.2pt
\begin{tabular}{lrrr}
\toprule
& time {\scriptsize [s] $\downarrow$} & speed {\scriptsize [sent./s] $\uparrow$} & space {\scriptsize [MiB] $\downarrow$} \\
\midrule
\multicolumn{4}{c}{\textsc{kyoto}} \\
MeCab & 28.53 & 62,495 & 40.52 \\
Vaporetto & 4.87 & 366,119 & 283.49 \\
Jagger (proposed) & 1.41 & 1,264,539 & 21.05 \\
\midrule
\multicolumn{4}{c}{\textsc{kwdlc}} \\
MeCab & 25.70 & 85,408 & 39.59 \\
Vaporreto & 4.87 & 366,119 & 283.49 \\
Jagger (proposed) & 1.13 & 1,942,477 & 20.16 \\
\bottomrule
\end{tabular}
\caption{Word segmentation efficiency.}\label{tab:seg}
\end{table}

\begin{table}[t]
\centering
\small
\tabcolsep4.8pt
\begin{tabular}{lccc}
\toprule
  & seg & top (level 1) & all (levels 1-4)\\
\midrule
\multicolumn{4}{c}{training: \textsc{kwdlc} $\rightarrow$ test: \textsc{kyoto}} \\
MeCab       & 97.90 & 96.56 & 94.82 \\
Vaporetto   &  95.76 & 93.81 & 91.31 \\
Jagger (proposed) & 97.25 & 95.42 & 93.30 \\
\midrule
\multicolumn{4}{c}{training: \textsc{kyoto} $\rightarrow$ test: \textsc{kwdlc}}\\
MeCab       &  97.78 & 96.02 & 94.48 \\
Vaporetto   & 97.05 & 95.15 &92.72 \\
Jagger (proposed) & 97.22 & 95.01 & 93.12 \\
\bottomrule
\end{tabular}
\caption{F$_1$ results for cross-domain evaluation.}\label{tab:kwdlc_kyoto}
\end{table}

\paragraph{Cross-domain evaluation} Lastly, Table~\ref{tab:kwdlc_kyoto} lists the results for cross-domain evaluation. Vaporetto's accuracy became much worse,
indicating that the classification-based method was prone to overfitting to the training domain. The proposed method enjoys the benefits of the dictionary and training data: it can change its behavior by adding not only dictionary entries but also patterns.

\section{Conclusions}
This study sought to improve the accuracy of speed-oriented, pattern-based methods for Japanese morphological analysis, rather than improving the speed of accuracy-oriented neural models. The proposed method extracts \textsc{pos}-augmented patterns from a morphological dictionary and annotated data. Experimental results on two standard datasets confirmed that this method achieves accuracy comparable to that of learning-based methods, with a very fast throughput of over 1,000,000 sentences per second on a laptop.

I plan to apply this approach to other languages and even to other \textsc{nlp} tasks by discretizing the continuous representations induced by neural models to obtain patterns. The source code is released with \textsc{gpl}, \textsc{lgpl}, and 2-clause \textsc{bsd} licenses.

\paragraph{Message to researchers} Because the accuracies on NLP benchmark datasets are becoming saturated with a larger foundation model, researchers may want to set diverse goals based on underrepresented metrics besides accuracy (\textit{e.g.}, efficiency). I hope that this study will initiate \textit{serious} research on speed-intensive approaches to \textsc{nlp} that can meet industry demands and enable researchers with limited computational resources to exert their ability.

\section{Limitations}
This evaluation had two limitations. First, although the method is not language-dependent, it was evaluated on a single language, Japanese. It would be worthwhile to evaluate the method on other languages to examine the approach's versatility. Second, the method uses dictionaries to obtain patterns. Although Japanese morphological analysis commonly uses dictionaries to perform lemmatization, it would be worthwhile to evaluate the method with only training data \yn{or dictionaries derived from text}.

Below, I discuss the current limitations for word segmentation, \textsc{pos} tagging, and lemmatization in detail.

\paragraph{Word segmentation} \yn{The proposed method's accuracy of word segmentation will depend on the target language's typological factors~\citep{shao-etal-2018-universal}, such as the character set size, lexicon size, and average word length. Among those factors, the character set size will especially matter because the current patterns mostly comprise surface strings and are likely to suffer from data sparseness. It will thus be valuable to evaluate the method on Chinese, which has a larger character set than Japanese. It will also be important to evaluate the method on languages with different typological factors from Japanese, such as Hebrew and Finnish. The training data size will not matter if the method is used to approximate some existing resource-efficient method via structure compilation~\citep{10.1145/1390156.1390231}.}


\paragraph{\textsc{pos} tagging} \yn{Compared to word segmentation, \textsc{pos} tagging requires more complex and abstract feature sets that are tailored for the target language and \textsc{pos} tag set~\citep{spoustova-etal-2009-semi}, which poses a challenge for the proposed method. The current pattern template is tailored for Japanese and the \textsc{juman} \textsc{pos} tag set; hence, for other languages and \textsc{pos} tag sets, a pattern template will need to be designed by referring to the feature templates of existing learning-based methods for the target language and \textsc{pos} tag set. Because the method jointly solves word segmentation and \textsc{pos} tagging in a left-to-right manner, patterns cannot leverage certain abstract features from posterior contexts of the target word (\textit{e.g.}, the next word's suffix). For application to other languages, it would be worthwhile to explore not only left-to-right processing but also right-to-left processing and a cascaded pipeline approach.}

\paragraph{Lemmatization} \yn{The approach here currently requires a morphological dictionary with lemmas or a fine-grained \textsc{pos} tag set that includes conjugation types and forms to perform lemmatization. Because lemma generation rules for other languages can be induced from lemma-annotated datasets~\citep{straka-2018-udpipe}, the method could be applied to other languages by using such lemma generation rules as the target labels for classification. Challenging target languages include morphologically rich languages such as Arabic and Czech.}

\section{Ethics Statement}
I am not aware of any specific social risks that this work directly creates or exacerbates. However, because morphological analysis is a core text processing function used in various \textsc{nlp} applications, those who attempt to abuse \textsc{nlp} applications may benefit from the proposed method's efficiency.

\section*{Acknowledgements}
This work was partially supported by JSPS KAKENHI Grant Number JP21H03494 and by JST, CREST Grant Number JPMJCR19A4, Japan. I thank Koichi Akabe for showing implementations of assigning \textsc{pos}s to unknown words in Vaporetto, Keiji Shinzato for his comments on an early draft of this paper, and Manabu Sassano for useful discussions on the future of speed-intensive \textsc{nlp}. Finally, I thank the anonymous reviewers for their encouraging comments on the paper's goal.

\bibliography{main}
\bibliographystyle{acl_natbib}

\appendix

\begin{table}[t]
\centering
\small
\tabcolsep3.3pt
\begin{tabular}{lrrr}
\toprule
\textsc{kyoto} & time {\scriptsize [s] $\downarrow$} & speed {\scriptsize [sent./s] $\uparrow$} & space {\scriptsize [MiB]  $\downarrow$} \\
\midrule
\multicolumn{4}{c}{w/ jumandic-5.1} \\
MeCab       & 26.83 & 66,455 & 55.81\\
Vibrato   & 12.47 & 142,983 & 97.75 \\
Vaporetto   & 15.14  & 117,767 & 658.80 \\
Jagger (proposed) &  1.77 & 1,007,344 & 26.39  \\
\midrule
\multicolumn{4}{c}{w/ jumandic-7.0} \\
MeCab       & 29.99 & 59,453 & 77.98 \\
Vibrato   & 16.01 & 111,367 & 164.20 \\
Vaporetto   & 16.93 & 105,316 &  828.85 \\
Jagger (proposed) &  1.83 & 974,316 & 35.09  \\
\bottomrule
\end{tabular}
\caption{Efficiency of morphological analysis on \textsc{kyoto}; results other than for Vibrato are from Table~\ref{tab:kc}.}\label{tab:kc_vib}
\end{table}

\section{Handling of Unknown Words}\label{app:unk}
Words that appear in neither the dictionary nor the training data matter in both the proposed method and search-based morphological analysis. Here, a common method~\citep{kudo-etal-2004-applying} was used to segment unknown words. Specifically, characters (and words) with the same character types, numbers, letters, or katakana were concatenated, with the concatenation restricted for katakana words when the total length of two katakana words exceeded a specific length (here, 18 bytes). The \textsc{pos} tags of concatenated unknown words were determined from a pattern based on the previous \textsc{pos} tag and the last concatenated word.

\section{Implementation Details}\label{app:impl}
Implementation techniques used in the existing efficient implementations of Japanese morphological analyzers were leveraged to implement Jagger. As in MeCab, memory-mapped \textsc{i/o} was adopted to reduce the memory footprint, and outputs are generated by referring to strings in the in-memory dictionary while avoiding dynamic memory allocation.
To maintain patterns, \yn{I used a character-wise, double-array trie that was adopted in Vaporetto and Vibrato.\footnote{\label{fn:vibrato}\url{https://github.com/daac-tools/vibrato}} To implement it, I modified an implementation of a byte-wise, double-array trie~\citep{yoshinaga-kitsuregawa-2014-self}, cedar.\footnote{\url{https://www.tkl.iis.u-tokyo.ac.jp/~ynaga/cedar/}} The character-wise, double-array trie uses \textsc{utf}-8 characters as atomic transition labels instead of \textsc{utf}-8 bytes}, which reduces the number of random accesses in traversing Japanese multi-byte characters. For the trie transition, \textsc{utf}-8 characters in the training data are counted to obtain cache-friendly, frequency-based \textsc{id}s for the \textsc{utf}-8 characters. These implementation tricks provided a total speed-up factor of at most two.

Note that block \textsc{i/o}, which outputs results with a fixed large size (256 KiB in these experiments), is crucial to maintain the method's very fast throughput when lengthy \textsc{pos} tags and lemmas are output.
The use of \texttt{strcpy} and \texttt{strlen} should be strictly avoided in formatting the output because they incur extra search for the terminal symbol \verb+\0+.

\begin{table}[t]
\centering
\small
\tabcolsep3.3pt
\begin{tabular}{lrrr}
\toprule
\textsc{kwdlc} & time {\scriptsize [s] $\downarrow$} & speed {\scriptsize [sent./s] $\uparrow$} & space {\scriptsize [MiB]  $\downarrow$} \\
\midrule
\multicolumn{4}{c}{w/ jumandic-5.1} \\
MeCab       &  23.83 & 92,110 & 53.88 \\
Vibrato   &  11.51 & 190,703 & 97.92 \\
Vaporetto   &  10.93  &  200,823 & 642.63 \\
Jagger (proposed) & 1.44 & 1,524,305 & 28.89 \\
\midrule
\multicolumn{4}{c}{w/ jumandic-7.0} \\
MeCab       & 26.90 & 81,598 & 76.38 \\
Vibrato   &  15.01 & 146,235 & 163.99\\
Vaporetto   & 12.55  & 174,900  & 842.40 \\
Jagger (proposed) & 1.46 & 1,503,424 & 40.22 \\
\bottomrule
\end{tabular}
\caption{Efficiency of morphological analysis on \textsc{kwdlc}; results other than for Vibrato are from Table~\ref{tab:kwdlc}.}\label{tab:kwdlc_vib}
\end{table}
\begin{table}[t]
\centering
\small
\tabcolsep3.2pt
\begin{tabular}{lrrr}
\toprule
& time {\scriptsize [s] $\downarrow$} & speed {\scriptsize [sent./s] $\uparrow$} & space {\scriptsize [MiB] $\downarrow$} \\
\midrule
\multicolumn{4}{c}{\textsc{kyoto}} \\
MeCab & 28.53 & 62,495 & 40.52 \\
Vibrato & 14.69 & 121,375 & 163.92 \\
Vaporetto & 4.87 & 366,119 & 283.49 \\
Jagger (proposed) & 1.41 & 1,264,539 & 21.05 \\
SentencePiece & 16.63 & 107,215 & 9.02 \\
\textsc{utf}-8 split & 0.31 & 5,751,612 & 1.55 \\
\midrule
\multicolumn{4}{c}{\textsc{kwdlc}} \\
MeCab & 25.70 & 85,408 & 39.59 \\
Vibrato & 13.94 & 157,460 & 164.30 \\
Vaporreto & 4.87 & 366,119 & 283.49 \\
Jagger (proposed) & 1.13 & 1,942,477 & 20.16 \\
SentencePiece & 14.54 & 150,962 & 9.05 \\
\textsc{utf}-8 split & 0.27 & 8,129,629 & 1.55 \\
\bottomrule
\end{tabular}
\caption{Efficiency of word segmentation (tokenization); some results are from Table~\ref{tab:seg}.}\label{tab:seg_vib}
\end{table}

\section{Comparison to Other Implementations}\label{app:vib}
\yn{I also compared Jagger with Vibrato (ver.~0.5.0),\footnoteref{fn:vibrato} which is a recent Rust reimplementation of MeCab by the developer of Vaporetto, and SentencePiece (ver.~0.1.99),\footnote{\url{https://github.com/google/sentencepiece}} which is an unsupervised text tokenizer for neural generation. SentencePiece was trained with the default options (vocabulary size of 8K) on the same training data.}

\yn{Tables~\ref{tab:kc_vib} and~\ref{tab:kwdlc_vib} summarize the efficiency of morphological analysis and Table~\ref{tab:seg_vib} summarizes the efficiency of word segmentation (tokenization) with the jumandic-7.0 dictionary. Although Vibrato is twice as fast as MeCab and shows comparable speed to Vaporetto for morphological analysis, Jagger is even faster and is more space-efficient than Vibrato. Jagger's throughput is on the same order as that of \textsc{utf-8} split, which simply looks at the first bytes (byte lengths) of \textsc{utf-8} characters to segment inputs into characters. Note that SentencePiece's small memory consumption is due to its small vocabulary size of 8K: it requires more memory for a larger vocabulary.}

\yn{Finally, it is noteworthy that the degree to which the processing speed is affected by the morphological dictionary's size varies from one implementation to another (Tables~\ref{tab:kc_vib} and~\ref{tab:kwdlc_vib}). Vibrato is the most affected by the dictionary size, whereas Jagger is the least affected.}



\end{document}